\def\year{2020}\relax
\documentclass[letterpaper]{article} 
\usepackage{aaai20}  
\usepackage{times}  
\usepackage{helvet} 
\usepackage{courier}  
\usepackage[hyphens]{url}  
\usepackage{graphicx} 
\usepackage{graphicx}  
\usepackage[utf8]{inputenc}         
\usepackage{acronym}                
\usepackage{todonotes}              
\usepackage{hyperref}               

\title{AI for Explaining Decisions in Multi-Agent Environments\thanks{This work was partly supported by Volkswagen Foundation.}}

\author{Sarit Kraus,\textsuperscript{\rm 1} Amos Azaria,\textsuperscript{\rm 2} Jelena Fiosina,\textsuperscript{\rm 3} Maike Greve,\textsuperscript{\rm 4} Noam Hazon,\textsuperscript{\rm 2}\\ 
\Large{\textbf{Lutz Kolbe,\textsuperscript{\rm 4} Tim-Benjamin Lembcke,\textsuperscript{\rm 4} Jörg P. Müller,\textsuperscript{\rm 3} Sören Schleibaum,\textsuperscript{\rm 3} Mark Vollrath\textsuperscript{\rm 5}}} \\ 
\textsuperscript{1}{Department of Computer Science, Bar-Ilan University, Israel} (email: sarit@cs.biu.ac.il)\\
\textsuperscript{2}{Department of Computer Science, Ariel University, Israel} \\
\textsuperscript{3}{Department of Informatics, TU Clausthal, Germany} \\
\textsuperscript{4}{Chair of Information Management, Georg-August-Universitat Göttingen, Germany} \\
\textsuperscript{5}{Chair of Engineering and Traffic Psychology, TU Braunschweig, Germany}
}

\newcommand{\joerg}[1]{\todo[inline, color=green!20]{#1}}

\newacro{mas}[MAS]{multi-agent system}
\newacro{xMASE}[xMASE]{Explainable decisions in  Multi-Agent Environments}
\newacro{XAI}[XAI]{Explainable AI}

\begin{document}

\maketitle

\begin{abstract}
Explanation is necessary for humans to understand and accept decisions made by an AI system when the system's goal is known. It is even more important when the AI system makes decisions in multi-agent environments where the human does not know the systems' goals since they may depend on other agents' preferences. In such situations, explanations should aim to increase user satisfaction, taking into account the system's decision, the user's and the other agents' preferences, the environment settings and properties such as fairness, envy and privacy.
Generating explanations that will increase user satisfaction is very challenging; to this end, we propose a new research direction: \ac{xMASE}.  We then review the state of the art and discuss research directions towards efficient methodologies and algorithms for generating explanations that will increase users' satisfaction from AI system's decisions in multi-agent environments. 
\end{abstract}

\section{Introduction}

Many AI systems need to make decisions in multi-agent environments where the agents, including people and robots, have possibly conflicting preferences. The system should balance between these preferences when making decisions regarding all agents. Such systems include, e.g., a scheduling algorithm assigning teachers to classes, or a ridesharing application proposing joint rides to people. 
In such situations, the global decisions made by the system may not adhere to all people's preferences: a decision may make some people unhappy. Providing explanations about the system's decision may increase people's satisfaction \cite{bradley2009dealing}, and maintain acceptability of the AI system \shortcite{miller2018explanation}. The EU General Data Protection Regulation introduces a right of explanation \cite{Goodman2016} for citizens to obtain “meaningful information of the logic involved” for automated decisions.

\ac{XAI} has recently been studied extensively~\cite{core2006building,carvalho2019machine,rosenfeld2019explainability}, 
mainly focusing on finding ways of explaining to a user a decision made by an AI system, e.g., a classification choice made by a neural network. That is, explanations are usually given for black-box algorithms, which aim at maximizing a well-agreed upon function (e.g., maximizing accuracy, minimizing loss). Previous work has mainly focused on increasing users' trust in the black-box AI system. However, we believe that providing the users with explanations is even more important in multi-agent environments, when even the maximization function is not clear to the (human) agents. In such situations, explanations should aim to increase user satisfaction, taking into account properties such as fairness, envy and privacy. Hence, we propose a new research direction of \acf{xMASE}.  

For example, in the ridesharing domain, an AI system may suggest to a customer Bob to share a taxi with Alice in a ride that will take 30 minutes.
The taxi will first drop Alice off and it will cost Bob \$25. Bob may be upset that the taxi drops Alice off first. An explanation could be that 
Alice's destination is on Bob's route and will only add 5 minutes to his trip. The system can also say that Alice will pay \$30, or that dropping Bob off first will add 15 minutes to Alice's trip, or that Alice teaches at 8 am and will be late if Bob is dropped first. Bob can also be told that sharing a taxi will save him \$10.

Generating explanations in multi-agent environments is even more challenging than providing explanations in other settings, e.g., for classification results produced by deep learning algorithms.
In addition to identifying the technical reasons that led to the decision, there is a need to convey the preferences of the agents that were involved. It is necessary to decide what to reveal from other agents' preferences in order to increase the user's satisfaction, taking the privacy of other agents into account, and how these preferences led to the final decision. It should also refer to issues such as fairness. The influence of the explanation on user satisfaction changes from one user to the next; therefore, personalized explanations are beneficial \cite{Lakkaraju2019FaithfulAC,bradley2009dealing}.
Given that the task is very challenging, we propose to use AI tools, and in particular machine learning to generate the personalized explanations that will maximize user satisfaction, while aiming to consider system-level societal welfare aspects. 
In order to use machine-learning methods for generating explanations in our context, it will be necessary to collect data about human satisfaction from decision-making when various types of explanations are given in different contexts. Furthermore, the evaluation of the proposed methods must also involve humans participating in multi-agent environments. 

Most algorithms that provide explanations on AI systems take an engineering approach, which does not involve running experiments with people. Papenmeier et al. \shortcite{papenmeier2019} showed that the presence of such explanations did not increase the subjects' self-reported trust. Miller \shortcite{miller2018explanation} argues that explanations have been studied extensively in psychology and their findings should be used when designing explanations for AI systems. One of his main points is that an explanation should be sensitive to context. We fully agree that for both
\ac{XAI} and \ac{xMASE} context should be taken into consideration when generating explanations, but in \acp{mas} context includes other agents' preferences, and fairness of the decision as an important factor. 

\begin{figure}
    \centering
    \includegraphics[width=8.5cm]{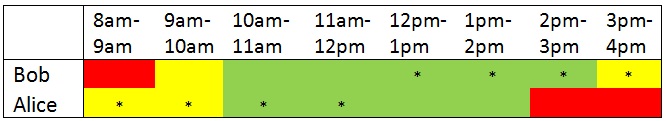}
        \caption{Scheduling example assigning teachers to classes, depicting Alice and Bob's preferences (green=preferred, yellow=feasible, red=impossible) and assignments (*).}
    \label{fig:AliceBob}
\end{figure}

For example, consider a scheduling algorithm that assigns teachers to classes. Suppose Bob and Alice each teach in a specific classroom on a specific day; each needs to teach for 4 hours (but they cannot teach in parallel).
Bob prefers to teach between 10am and 3pm, but can also teach from 9am-10am and 3pm-4pm. He cannot teach between 8am and 9am (a strong constraint) or after 4pm. Alice prefers to teach between 10am-2pm, has a hard constraint not allowing her to teach after 2pm, but can teach in the mornings, 8am-10am. Suppose the algorithm assigned Bob to 12pm-4pm and Alice to 8am-12pm (see Figure \ref{fig:AliceBob}). Which explanations should the system provide for Alice whose assignment violates her soft constraint but not her hard constraints? Which explanations should the system give a teacher for an assignment violating his or her hard constraints? 
Suppose Alice and Bob are friends. Should the system tell Bob that he teaches between 3pm and 4pm (violating his soft constraint) because Alice has a strong constraint at this time? What if Alice and Bob are new to the school and do not know one another; should such an explanation be provided?
Alternatively, could a merely graphical description of the relevant information (as in Figure~\ref{fig:AliceBob}) suffice as an explanation? 

There are many challenging research questions to be studied in order to provide users in \acp{mas} with explanations that will increase their satisfaction:

\begin{description}
\item[Algorithms for Explanation Generation:] The develop\-ment of efficient algorithms that will generate the explanations, preferably in real time, is very challenging. First, there is a need to be able to form good explanations. Then, there is a need to decide which explanation to present (if at all) and when and how to present it.  
\item[User modeling for increasing satisfaction:] Most important is to be able to model the users. It is, of course, necessary to identify the users' preferences in order to make good decisions. However, for generating good explanations it is important (beyond obtaining the users' preferences) to also model their attitudes toward different explanations. That is, to be able to predict how an explanation will influence users' satisfaction. 
\item[Interactive explanations:]
To address the difficulties in choosing the specific explanation to a specific user, we propose  to also study interactive explanations, which are provided to the user through a dialogue between the AI system and the user \cite{miller2018explanation}. By asking questions and expressing his or her concerns, the user can direct the system towards generating good explanations. However, conducting meaningful dialogues with a user for \ac{xMASE} adds difficulties to the process. It has some similarities to automated systems that argue with people, a research area that still has many open questions \cite{rosenfeld2016providing,rosenfeld2016strategical}.
\item[Understanding System Decision-making:] The decision of an \ac{xMASE} system depends on many parameters associated with several agents, making the problem even more challenging than in \ac{XAI}. Only some of the technical reasons that led to the decision are relevant to a given user; as the number of agents increases, the non-relevant information increases, too, and it is hard to identify the relevant parts. Other explanations that could increase user satisfaction concern the environment. If the AI system made a decision based on some knowledge about the environment, but this knowledge is not available to the user, presenting it to the user could be useful. E.g., if Bob believes that there are many taxis available at 8am, but the system proposed him to share a taxi with Alice since it knows that there are very few taxis available, giving this information to Bob might increase his satisfaction. 
\item[ Long-term relationships:] When the AI system interacts repeatedly with the same users, interesting research questions may arise. The learning phase of the preferences and satisfaction models can be personalized, but more importantly the explanation generated should take long-term satisfaction into consideration. 
\item[Ethics and Privacy:] These issues must be considered when presenting explanations in multi-agent systems. As in other situations, one needs to regard issues such as the truthfulness of the explanations and the ethical aspect of concealing some of the information. Moreover, privacy is a major concern in \ac{xMASE}, since there is also a need to consider the revelation of information and preferences of one agent to another when providing an explanation.
\item[Open source code and public datasets:] Open source environments that will allow researchers to develop and evaluate their explanation algorithms can enhance the research in \ac{xMASE}. It is also very useful to start collecting labeled datasets for \ac{xMASE}. 
\end{description}
We will discuss three of these  directions in more detail after surveying the related work to \ac{xMASE}.

\section{State-of-the-art}
In recent years there has been intense research on  design techniques for making AI methods explainable, interpretable, and transparent for developers and users \cite{carvalho2019machine}.
The basic idea of \ac{XAI} methods is to try to explain black-box model behaviour (while \ac{xMASE} is needed even in settings where traditional white-box optimization is used). Methods for \ac{XAI} have been developed including locally interpretable model-agnostic explanations for Bayesian predictive models \cite{Peltola2018LocalIM} and for convolutional neural networks \cite{Mishra2017LocalIM}, visualization techniques (Grad-CAM) for CNNs \cite{Selvaraju17}, or black box explanations through transparent approximations \cite{Lakkaraju17}. Hybrid models use explicit symbolic representations in conjunction with black-box techniques~\cite{hu2016harnessing,choi2019relative}. \ac{XAI} was also expanded beyond the classical domains. For example, Fox et al.~\shortcite{fox2017explainable} introduces explainable planning systems. Ludwig et al.~\shortcite{ludwig2018explaining} and {\v{C}}yras et al.~\shortcite{vcyras2019argumentation} investigated the explainability of a scheduling system over tasks. We note that both planning and scheduling of tasks to machines are not multi-agent environments as we define them, since they actually consist of a single agent, and therefore do not belong to \ac{xMASE}. 
The explainability of deep reinforcement learning was also investigated~\cite{Lee19}.
All of these techniques were developed for \ac{XAI},
and they can serve as input for \ac{xMASE} that needs to choose the suitable explanation for any given scenario and to each agent. Moreover, in many \ac{XAI} approaches the evaluation is complicated due to missing standards~\cite{pedreschi2019meaningful1}, the social nature of explanations is ignored~\cite{miller2018explanation}, and the explanations are not evaluated with humans. All of these evaluation criteria are essential in \ac{xMASE}.

Indeed, there are some works which provide evaluations of a \ac{XAI} approach with humans. 
Lakkaraju et al.~\shortcite{Lakkaraju2019FaithfulAC} propose a new form of explanations designed to help end users (e.g., decision-makers such as judges, doctors) to gain a deeper understanding of the models' behaviour. 
Doshi-Velez and Kim \shortcite{doshi-velez.2018} claim that researchers evaluating explainability should differentiate between evaluation of explanations with humans, experts in the evaluated fields, and a formal evaluation without human subjects. 
Wolf et al. \shortcite{wolf2019} point out requirements for explanations and state that these should be influenced by the users, the applications and the deployment context. 
They claim that three different types of explanations are needed for integrating \ac{XAI} approaches into real world applications. Firstly, an application should explain its behavior; Secondly, the impact of the interaction of users with the  applications should be explained, and lastly, a user seeks explanations describing how the applications' output integrates in the overall process. These ideas are also some of the basic ingredients of a successful \ac{xMASE}.

One unique aspect of \ac{xMASE} is that the explanations should be chosen such that they will increase the user's satisfaction. Previous works have shown that explanations in general have an impact on user satisfaction/acceptance. Herlocker et. al. \shortcite{herlocker2000explaining} showed that providing explanations
for automated collaborative filtering (ACF) recommendations can improve the acceptance of ACF systems.
More relevant to us is \cite{kleinerman2018providing} showing that explanations are beneficial also in reciprocal recommendation systems (e.g., dating) where the preferences of the other agent are taken into account when making recommendations, and possibly when generating explanations. In such settings reciprocal explanations are preferable.  
Putnam and Conati \shortcite{putnam2019exploring} conducted a user study on benefits from  explanations in intelligent tutoring systems. Their results indicate a positive sentiment towards wanting explanations, but do not suggest any automatic system for their generation.
%
Levinger et al. \shortcite{levinger2018human} study maximizing human satisfaction in multi-agent systems, but do not use any form of explanation. They present an optimization algorithm that maximizes the overall human satisfaction according to the learned model. They also show that, when aiming at maximizing human satisfaction, learning accurate models of human satisfaction is more important than improving the optimization algorithm.

\section{Research directions for \ac{xMASE}}

The development of AI-based tools that provide the right explanations to the right users at the 
right time to increase user satisfaction in \acp{mas} is very challenging. We now discuss three of the above research directions in more detail.

\subsection{Generation of explanations to increase satisfaction} 
The development of efficient algorithms that will generate the explanations, preferably in real time, is very important. 
We propose a two stage procedure: first, a set of possible explanations will be created and then the one that best suits the specific user at the specific settings will be selected. Both stages can be done using machine learning or any other decision-making procedures based on real user input.

If the AI decision is made using neural networks (e.g., \cite{rosemarin2019emergency,li2019efficient}) then XAI methods can be used to identify important features that led to the decision
\cite{Shrikumar17,Bach15}. These methods should be adapted to \ac{xMASE}-related problems \cite{Lee19,Selvaraju17}.
Then, there is a need to identify which of these features are relevant to a specific user. Given these features, the preferences of other agents that are relevant should be identified and any relevant statements that touch upon important concepts such as  fairness should be generated.
Using these features, preferences and concepts, several explanations could be generated using subsets of them. Next, the explanations with the relevant preferences and environment settings could be entered into a network that will estimate the influence level of each explanation on the user's satisfaction. Finally, the chosen subset should be transferred into a textual message and sent to the user. Personalization could also be used in this final step.

If the AI decision-making is not done using machine learning methods, but rather, e.g., using inferences, and data is not available, it will be interesting to study how methods developed for explaining inferences (e.g.,  \cite{pino2003preferences}) could be used for \ac{xMASE}.

If the AI decision-maker is a scheduling tool, it can provide a set of constraints that lead to the proposed schedule \cite{ludwig2018explaining}. In \ac{xMASE} there will be a need to identify the relevant constraints and to generalize  statements related to other agents' preferences, and the general system constraints that are driven by other concepts such as fairness. Then again, we can use user satisfaction models (e.g., represented by a neural network) to choose the best constraints and generalized statements to be presented.

Complementing  the algorithms and methodologies described so far, an interesting research direction is the automated generation of graphical explanations for xMASE, enabling compact summarization of large amounts of information. 
\subsection{User modeling for increasing satisfaction}
There are many methods for user modeling and preference elicitation \cite{rosenfeld2018predicting,anselmi2018comparison}, but we have not found a study on preference elicitation with respect to explanations and their role in improving satisfaction. One of the main challenges is that user  satisfaction from an explanation of a given decision strongly depends on the actual decision, the other agents, the environment and the user's beliefs. For example, it is obvious that, in our ridesharing example, an explanation to Alice of the form ``This shared ride will save you \$10" is a ``good" explanation, but of course if in the specific ride Alice will save time but not money, the explanation is useless. Similarly, telling Bob that ``There is no private taxi available in your area now" may be a good explanation for a ridesharing suggestion, given that it is true (assuming we provide only true explanations) and Bob does not know it. Thus collecting data on the influence of an explanation on the user's satisfaction must be done in the context of the specific decision it explains and the environment setting. This makes data collection very challenging.

Collection of data can be done either using fictitious decisions, their explanations and the \ac{mas} environment setting or, much harder to accomplish, in actual settings or at least in simulations. 
The users can express their preferences on how much they like the explanations. We can use this data to build a generalized model of users' preferences toward explanations. However, this model will not provide us with the explanations that increase the user's satisfaction. Here we will need to let the user express his or her level of  satisfaction from a given decision with different variants of explanations and without explanations, and try to build a model that measures the users' satisfaction from the decision.

One of the challenges is to identify features of an explanation. We may consider using the explanation as text, but we believe that there is a need to find additional features that have to do with the relationship of the explanation with the user's preferences and the environment. 

When the AI system interacts repeatedly with the same users, interesting research questions may arise. The learning phase of the preferences and satisfaction models can be personalized, but more importantly the explanation generated should take long-term satisfaction into consideration.
Furthermore, we propose to consider, when interacting with the user, using reinforcement learning to improve the user's model of overtime in a guided way.
\subsection{Interactive explanations}   
Human verbal explanations are essentially interactive \cite{cawsey1993planning}.
Recently, there have been a few attempts to consider models for interactive explanations to XAI \cite{madumal2018towards}
and to value-based agents
\cite{liao2018representation}, but no system was developed. Interactive explanations can be viewed as argumentation dialogues. It was shown to be beneficial to model the interaction as POMDP where the uncertainty is about the user's beliefs \cite{rosenfeld2016strategical}.  Using this approach for xMASE there is a need to continuously estimate the user's beliefs and sentiment toward the AI system's decision, and to predict how a given explanation statement will modify the user's beliefs and influence its attitude toward the decision.

Other techniques for general dialogue systems or for interactive learning dialogues could be considered \cite{chen2017survey}. The open questions are how to (1) use the features that led the AI system to make a decision to form possible responds, and (2) devise user models of preferences and satisfaction to generate the right response in the dialogue. 
Regardless of the techniques, any developed method should be evaluated via human experiments, which are still often missing in XAI.
\section{Conclusions}
In this paper we presented the xMASE challenge for explaining AI-based systems in multi-agent systems environments aiming at increasing user satisfaction. This challenge is extremely important for the success and acceptability of socio-technical applications such as ridesharing where people's preferences may conflict, but cooperation is beneficial. xMASE can build on top of XAI's recent progress, but many open questions should be addressed which are related to the other agents in the environment and the goal of increasing user's satisfaction. We propose to develop AI-based techniques toward addressing these challenges.
\bibliographystyle{aaai}
\bibliography{bluesky}

\end{document}